\documentclass[10pt,twocolumn,letterpaper]{article}

\usepackage{iccv}
\usepackage{times}
\usepackage{epsfig}
\usepackage{graphicx}
\usepackage{amsmath}
\usepackage{amssymb}
\usepackage{booktabs}
\usepackage{subcaption}
\usepackage{fixltx2e}

\usepackage[pagebackref=true,breaklinks=true,letterpaper=true,colorlinks,bookmarks=false]{hyperref}

 \iccvfinalcopy 

\def\iccvPaperID{12371} 

\begin{document}

\title{HAISTA-NET: Human Assisted Instance Segmentation Through Attention}

\author{Muhammed Korkmaz\\
Koc University\\
Istanbul, Turkey\\
{\tt\small mkorkmaz20@ku.edu.tr}
\and
T. Metin Sezgin\\
Koc University\\
Istanbul, Turkey\\
{\tt\small mtsezgin@ku.edu.tr}
}

\maketitle
\ificcvfinal\thispagestyle{empty}\fi

\begin{abstract}
Instance segmentation is a form of image detection which has a range of applications, such as object refinement, medical image analysis, and image/video editing, all of which demand a high degree of accuracy. However, this precision is often beyond the reach of what even state-of-the-art, fully automated instance segmentation algorithms can deliver. The performance gap becomes particularly prohibitive for small and complex objects. Practitioners typically resort to fully manual annotation, which can be a laborious process. In order to overcome this problem, we propose a novel approach to enable more precise predictions and generate higher-quality segmentation masks for high-curvature, complex and small-scale objects. Our human-assisted segmentation model, HAISTA-NET, augments the existing Strong Mask R-CNN network to incorporate human-specified partial boundaries. We also present a dataset of hand-drawn partial object boundaries, which we refer to as “human attention maps.” In addition, the Partial Sketch Object Boundaries (PSOB) dataset contains hand-drawn partial object boundaries which represent curvatures of an object’s ground truth mask with several pixels. Through extensive evaluation using the PSOB dataset, we show that HAISTA-NET outperforms state-of-the art methods such as Mask R-CNN, Strong Mask R-CNN, and Mask2Former, achieving respective increases of +36.7, +29.6, and +26.5 points in AP$\textsubscript{Mask}$ metrics for these three models. We hope that our novel approach will set a baseline for future human-aided deep learning models by combining fully automated and interactive instance segmentation architectures.
\end{abstract}
\begin{figure}[htbp]
\centering%
   \includegraphics[width=1\linewidth]{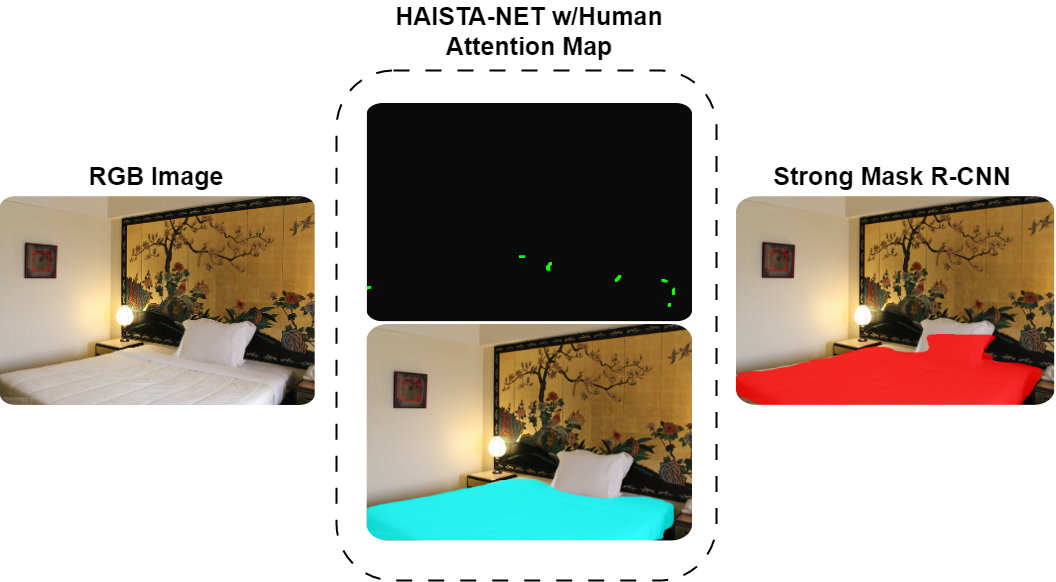}
   \caption{First glance of precise mask prediction by HAISTA-NET using human attention maps dataset (center image). HAISTA-NET outperforms the mask prediction of Strong Mask R-CNN (right) on high-curvature objects.}
   \label{fig:intro}
\end{figure}
\section{Introduction}
In recent years, demand has grown for deep-learning-based, fully automated instance segmentation models \cite{instance1,instance2,instance3,instance4,instance5,instance6,instance7,instance8,instance9}. Due to rapid progress in their development, these image detection tools have become the preferred method for applications such as object refinement \cite{refine}, medical image screening \cite{medical1,medical2}, and image/video editing \cite{video1,imageedit1}. However, they often perform poorly when objects are small-scale or have a pronounced 
curvature, causing inaccuracies in the instance segmentation mask, such as boundary outflows or under-covering \cite{boundary}. To improve mask precision, users may resort to manual annotation \cite{manualann1}, though this has several drawbacks including the added cost of time.

As a result, researchers have been studying alternative approaches to interactive segmentation \cite{int1, int2, int3, int4}, to remedy existing deficiencies in segmentation masks while reducing the time needed for manual annotation. One method involves deep-learning-based interactive segmentation models that use corrective action techniques for object mask retrieval \cite{int1, adobe}. Another proposed technique involves mask editing based on mouse clicks \cite{clickinteraction}, whereby the user helps to guide the automated model manually by detecting boundary outflows or under-covering.

With HAISTA-NET, we bring a new approach to instance segmentation by combining two main areas: fully automated instance segmentation, and interactive instance segmentation. With this proposed method, users convey the intended boundary by marking it manually. A deep learning network, which has previously been trained with such markings, uses this input to drive the instance segmentation process. Our novel approach eliminates mask errors by training the model with user inputs, to fix the mask where other models stumble and fail.

We also present our new Partial Sketch Object Boundaries (PSOB) dataset. This dataset has images, objects, and categories taken from LVIS \cite{lvis} and extended with additional hand-drawn sketches. It contains raster images, which we refer to as human attention maps, drawn by users with a few pixels (minimal input) of the high-curvature regions of the object boundary, where segmentation masks are most erroneous. For the PSOB dataset, 30 users annotated 18,677 objects of different scales and a varying number of high curvature sections.

HAISTA-NET architecture uses human attention maps both for the training phase and for running inference. Our model can be easily integrated with various deep learning-based segmentation and detection models (see \ref{sec:Network Architecture}). HAISTA-NET uses a Strong Mask R-CNN baseline \cite{instance1, simplecopy}. In our model we removed the first convolution layer of the Mask R-CNN \cite{instance1} backbone and added a new one that is fed by a combination of the Human Attention Map and a three-channel RGB image. As a result of our experiments, the HAISTA-NET architecture outperforms the Strong Mask R-CNN baseline with a more precise mask for high-curvature and/or small-scale objects (Figure ~\ref{fig:intro}).

\begin{figure*}
\centering%
\includegraphics[width=0.8\linewidth]{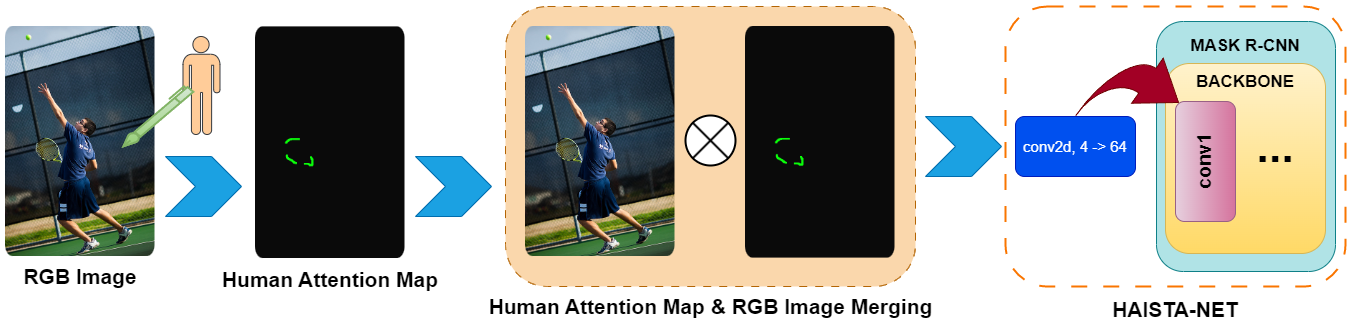}
   \caption{ \textbf{Method Outline.} Users draw a couple of pixels according to their attention to the object. Then, a Human Attention Map is generated to concatenate with the RGB image to feed the model. We denote the concatenate operator as $\otimes$.}
\label{fig:outline}
\end{figure*}
We also developed a user-friendly interface that allows users to interact with objects to create or edit human attention maps using partial strokes \cite{sketch}. By using this interface, users can either directly annotate images without prior knowledge of the segmentation results or perform the annotation after viewing outputs from the Strong Mask R-CNN.

When using the PSOB dataset with human attention maps, our model achieves more accurate results than existing state-of-the-art models. Evaluations conducted with the PSOB dataset show that our model achieves a performance that is 36.7 points better than Mask R-CNN, according to the AP$\textsubscript{Mask}$ metric, and +29.6 points compared with Strong Mask R-CNN. Using the AP$\textsubscript{Bbox}$ metric, our model also demonstrates an increase of +33.5 and +31.1 points, respectively, versus Mask R-CNN and Strong Mask R-CNN (see Section ~\ref{sec:MainResults}). Moreover, HAISTA-NET achieves a +26.5-point increase in AP$\textsubscript{Mask}$ versus Mask2Former \cite{instance5}, which is the current state-of-the-art model for instance segmentation on the COCO \cite{coco} dataset. 

Our main contributions can be summarized as follows:
\begin{itemize}
\item We have developed HAISTA-NET based on the Mask R-CNN architecture so that three-channel RGB images can be combined with a human attention map.
\item We propose a sketch-based representation of user-defined, high-curvature sections of objects of interest, called the human attention map.
\item We present the Partial Sketch Object Boundaries (PSOB) dataset, which will be a valuable asset in driving further research in user-assisted instance segmentation.
\item We propose the Adaptive Curvature Number Detector (ACND) for detecting and classifying the curvatures of segmentation masks.
\item Applying multiple factor analysis, we have reported our model results for objects according to different scales, different curvature numbers, and different user drawing characteristics.
\item Our user study also provides analysis of the cost of sketch-based manual annotations in LVIS dataset quality.
\end{itemize}
\section{Related Work}
Instance segmentation methods can be divided into two main categories: fully automated instance segmentation \cite{instance1,instance2,instance3,instance4,instance5,instance6,instance7,instance8,instance9}, and interactive instance segmentation \cite{int1, int2, int3, int4, video1, imageedit1, adobe}. The Strong Mask R-CNN \cite{instance1, simplecopy} tool is one of the state-of-the-art models available that achieves a high degree of precision. This architecture extends from the Faster R-CNN \cite{fasterrcnn} software and sets a baseline for many deep learning models with its simplicity and ease of implementation. With the contributions of Ghiasi \etal \cite{simplecopy}, the simple copy-paste method has improved the performance of the basic Mask R-CNN architecture and transformed it into the Strong Mask R-CNN model. Strong Mask R-CNN, our baseline, became prominent with its practical use, namely by obtaining high precision masks with fast forward training and low memory usage.

Transformer-based architectures \cite{transformer1, transformer2, transformer3} are increasing in popularity due to their high image feature learning and attention mechanisms. For example, Mask2Former \cite{instance5} architecture uses transformers in place of the conventional RPN backbone of Faster R-CNN, and surpasses both standard Mask R-CNN \cite{instance1} and Strong Mask R-CNN \cite{instance1, simplecopy} in results with its transformer-based structural design. These new-generation transformers are replacements of CNN-based traditional architectures. Yet such models still have a high memory and time cost \cite{costmem}, not only for training but also for inference. Also, they tend to under-perform in fine-grained applications such as medical image analysis \cite{medicalimage2}.

To avoid segmentation failures in object boundaries, some researchers use time-costly manual annotation techniques. Interactive segmentation is another alternative that has emerged recently and promises superior results to manual annotation. Some studies suggest using mouse clicks or free-style painting \cite{scribble1} to correct the boundaries of segmentation masks, assuming this reduces annotation time.

According to Benenson \etal \cite{int1}, the point and click-based \cite{objectselection} method is an effective technique to improve mask precision rates. Using mouse clicks requires two types of marking for redefining object boundaries. If the estimated segmentation mask overflows the actual boundary of the object, the user sets negative markers to ignore those segments. If the boundary under-covers the ground truth, the user sets the targeted distance by placing positive markers. However, the mouse click technique may fail for small-scale objects as it requires a high level of detail to adjust in bounding box limits.

In this study, instead of using time-consuming mouse clicks to edit faulty masks, we advocate partial marking of the object boundary prior to instance segmentation, and then using these marks to aid segmentation. The boundary is marked specifically at problematic high-curvature regions where the segmentation fails the most.
\section{Proposed Approach}
We propose a methodology to generate more precise segmentation masks by integrating human attention maps with fully automated segmentation models \cite{instance1, simplecopy}. Our study includes the following steps: a review of dataset collection, data annotation, selection of input representation techniques, conversion of input representations into human attention maps, demonstration of network architecture, fine tuning of training parameters, and running inference.
\subsection{Partial Sketch Object Boundaries Dataset}
\label{sec:PartialSketchObjectBoundariesDataset}
When first creating our dataset and deciding on which annotation technique to use we explored the failure conditions of different instance segmentation models. First, we trained a Strong Mask R-CNN \cite{instance1, simplecopy} model with the LVIS dataset to investigate segmentation mask failures. Following this training, we selected 3,070 test images from the LVIS \cite{lvis} dataset to run inference in order to determine object mask failures. The object mask prediction was not precise in the following conditions:
\begin{itemize}
\item If the object scale is small, with an \(area < 32^2\).
\item If the object is medium-scale \(32^2 < area < 96^2\) or large-scale \(area > 96^2\) and has a high degree of curvature or number of curves.
\item If the object is located behind another object, causing an occlusion problem.
\item If the object’s shape and curvature differ in test images from training images.
\end{itemize}
\begin{figure}[htbp]
\centering%
   \includegraphics[width=0.8\linewidth]{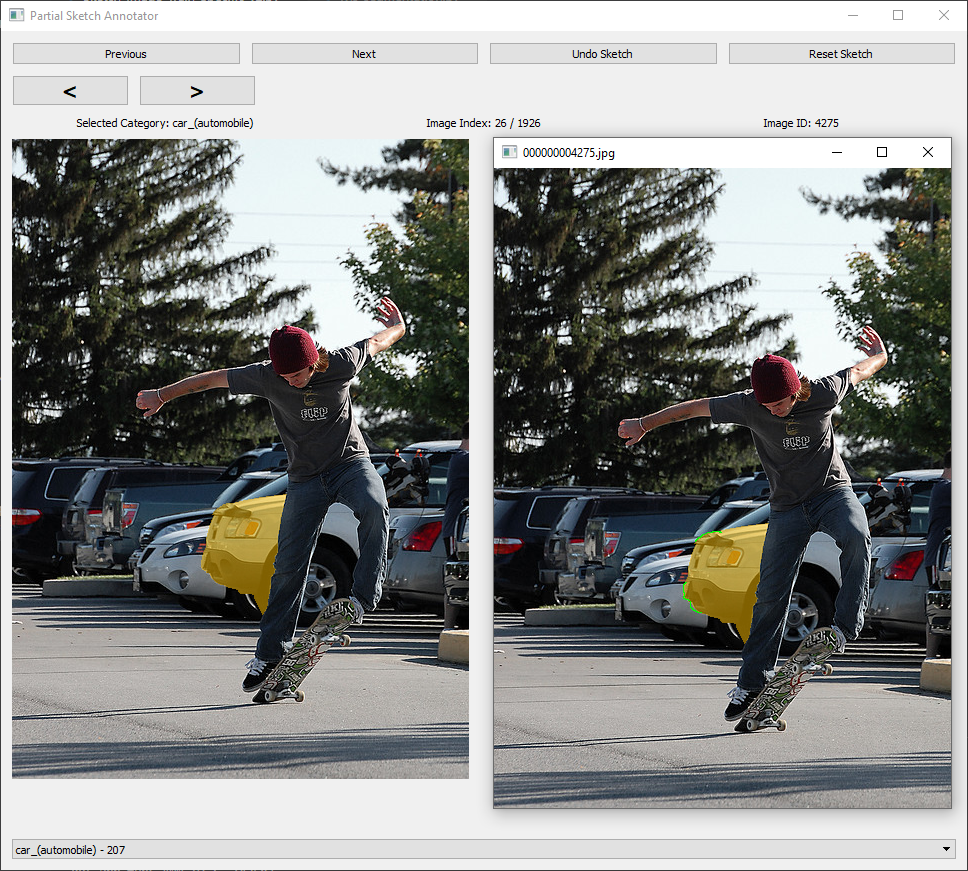}
   \caption{\textbf{Interactive interface.} Users can interact with the target object via this tool.}
\label{fig:interface}
\end{figure}
Regarding the analysis results, we propose a minimal sketch-based technique to extend the LVIS dataset with user input. Our newly constituted dataset, PSOB, contains hand-drawn partial object boundaries in high-curvature points (see Section ~\ref{sec:AdaptiveObjectCurvatureDetector}) for 115 categories of objects taken from the LVIS dataset. Thirty users annotated data, using hand-drawn partial sketches to represent object high-curvature points. Our aim was to minimize human effort while interacting with the object. For the data distribution to be reasonable, we annotated 10,450 training objects, 4,109 validation objects, and 4,118 test objects. For experimental reasons, we also recorded the object’s total sketch time [sec], time for each stroke [sec], and the pixel-wise stroke’s corresponding values in the x/y coordinate frame.

The important factor of this dataset is human attention
points, which represent user inputs to form human attention maps. To obtain these points, some parts of the boundaries of the objects of interest are sketched by users. These sketched points represent human assistance in our work. We identify this assistance according to the following rules:
\begin{itemize}
\item If sketch input covers less than 25\% of the object’s boundary, human assistance is designated as \textbf{minor assistance}.
\item If sketch input covers between 25\% and 50\% of the object’s boundary, human assistance is designated as \textbf{medium assistance}.
\item If sketch input covers more than 50\% of the object boundary, human assistance is designated as \textbf{major assistance}.
\end{itemize}
\subsection{Adaptive Object Curvature Detector}
\label{sec:AdaptiveObjectCurvatureDetector}
Object masks consist of polygons having different angles between line segments. Angles of intersecting line segments constitute a high-curvature or low-curvature point depending on their degree. We use the Ramer-Douglas-Peucker (RDP) \cite{rdp} algorithm to refine curvature points for each object boundaries.

Object masks consist of polygons with different angles between their line segments. Angles of intersecting line segments constitute a high-curvature or low-curvature point depending on their degree. We use the Ramer-Douglas Peucker (RDP) \cite{rdp}  algorithm to refine curvature points for each object boundary.

The standard RDP algorithm uses a static epsilon value to simplify the number of the object’s corners. Designated static RDP epsilon values require parameter tuning for every scale of the object. In order to fine tune this value, we calculate the perimeter of the polygon from the coordinates and set the epsilon value to 3\% of the perimeter. This method returns more accurate results while calculating the number of curvature points for objects from different scales, compared with classical RDP. We use the following formula, where $\epsilon$ is adaptive epsilon for RDP, ${n}$ is the line segment count, and $\mathcal{I}$ represents line segments:
\begin{equation}
\epsilon = {0.03} \cdot \sum_{k=1}^{n}\sqrt{(\mathcal{I}_{k_{x_2}}-\mathcal{I}_{k_{x_1}})^{2}+(\mathcal{I}_{k_{y_2}}-\mathcal{I}_{k_{y_1}})^{2}}
\label{eq:epsilon}
\end{equation}
After obtaining the results of the adaptive object curvature detector (eq. ~\ref{eq:epsilon}), we classify the objects according to their number of curvature points by the using following rules:
\begin{itemize}
\item If an object’s number of curvature points is less than six, an object’s curvature type is designated as \textbf{low curvature}.
\item If an object’s number of curvature points is between six and 10, its curvature type is designated as \textbf{medium curvature}.
\item If an object’s number of curvature points is greater than 10, it is designated as a \textbf{high curvature} type.
\end{itemize}
\subsection{Representation of Human Attention Map}
\label{sec:RepresentationofHumanAttentionMap}
As briefly described in previous sections, high-curvature points on the objects of interest are partially annotated by human users in order to create human attention maps. The x and y pixel coordinates of these user strokes are used to form binary images with the same dimensions as input images. These binary images are representations of the human attention map and are later supplied to the fourth channel of the first convolution layer, where the other three channels are reserved for the original RGB image. Attention points that are annotated by human users as high curvature points are set to the pixel value of 255 while other regions are left initially as zero. This results in a sparse representation where most of the image is zero while a few pixels at the location of interest are 255. During the training phase, random initialization of the fourth channel weights were used.

During the preliminary analysis stage, the effects of representing unrelated regions with various forms were investigated. Since, during back-propagation, randomly multiplying the initialized weights of the fourth channel with zeros prevents optimization of these weights, alternative strategies were tested, such as using a small fixed value or using a randomly generated small number instead of zeros.

The results show an insignificant difference in performance among the three methods proposed for representing unrelated regions, namely: using zeros, using a small fixed pixel value ($\mathcal{P}$) such as 10, and using a randomly generated number. Because of this, using a fixed pixel value of 10 for regions outside of attention points was selected for the rest of the study (eq.~\ref{eq:attention}).
\begin{equation}
\label{eq:attention}
    \mathcal{P}= 
\begin{cases}

    255,             & \text{if location is attention point} \\
    10,              & \text{otherwise}
\end{cases}
\end{equation}
\begin{figure*}
\centering%
   \includegraphics[width=1\linewidth]{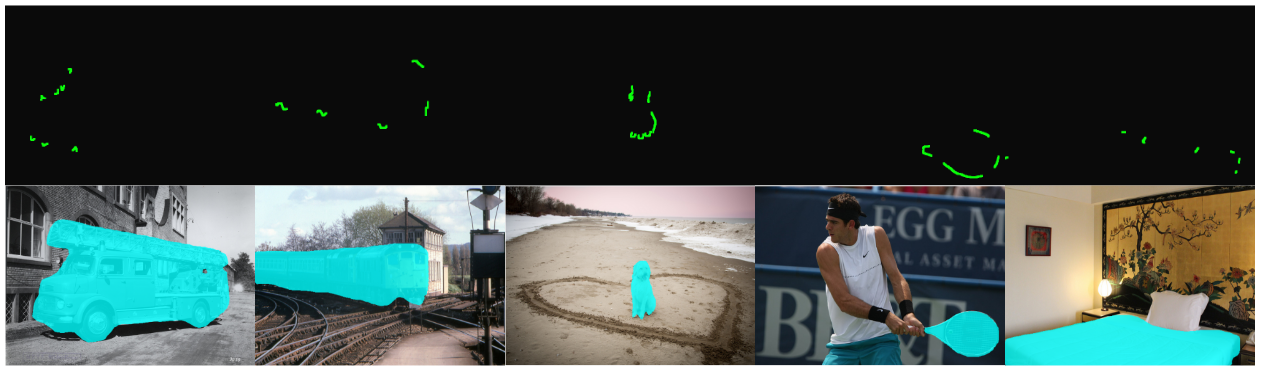}
   \caption{\textbf{Visualization of the Predictions of HAISTA-NET.} We present images with different scales, curvature numbers, and hand-drawing-based assistance types.}
\label{fig:predictions}
\end{figure*}
\subsection{Network Architecture}
\label{sec:Network Architecture}
HAISTA-NET adds a head branch to the Mask R-CNN \cite{instance1} architecture. We remove the first convolution from the backbone of Mask R-CNN to set our 2D convolution. We concatenate the 3-channel RGB image with our human attention map to form 4-channel image inputs (Figure ~\ref{fig:outline}). We set the new convolution that overrides the first convolution of the current backbone with an input channel parameter equal to 4 and an output channel parameter equal to 64, as represented in standard Res-Net \cite{resnet} architecture. The new convolution kernel size is 7x7, the stride is 2x2, and the padding is 3x3. While adopting the new weights of the 4-channel convolution, for the first three channels, instead of randomly initializing weights, we use pre-trained weights from the existing 3-channel convolution. Because the human attention map channel is new, we randomly initialize the weights between a range of (0, 0.001). As we described earlier, Mask R-CNN \cite{instance1} architecture extends from Faster R-CNN \cite{fasterrcnn}. The region proposal network backbone of Faster R-CNN uses data transformation parameters of mean and standard deviation vectors to adopt new input values when the channel number increases or decreases. Therefore, we need to adjust the final channel’s mean and standard deviation vectors. We set the mean of the final channel as 0.5. The mean vector of first three channels is (0.485, 0.456, 0.406).

Additionally, we set the standard deviation of the final channel to 0.2, where the standard deviation vector for the first three channels is (0.229, 0.224, 0.225). We evaluate Res-Net \cite{resnet} and Res-NeXt \cite{resnext} architectures of depth 50, 101, and 152 layers with feature pyramid networks (FPN) \cite{fpn}. We achieved the best results with Res-NeXt-101-FPN.
\subsection{Data Augmentation}
\label{sec:DataAugmentation}
In order to develop our enhanced model, we performed different training techniques \cite{strategy}. Firstly, we flipped the image randomly and assigned the probability value of being flipped to 0.5. As a result of this, the sample diversity increased. Secondly, we used the large-scale jitter augmentation technique to randomly resize the image and image bounding box. We resized the image and boundary within the scale range (0.1, 2). The transform of the target size was 1024x1024. We also used bilinear interpolation as a parameter. Finally, we used the fixed-size crop to scale the image to the target size. Before feeding the model with our dataset, we used the simple copy-paste \cite{simplecopy} data augmentation method to create images from rare categories. We took object ground truth annotations from two images, then created a new image using these annotations to reproduce samples. Finally, we compared the original 3-channel Mask R-CNN and our own model regarding the effectiveness of this simple copy-paste data augmentation method.

The results show that Mask R-CNN achieves $\approx$7 AP improvements while we achieve less than $\approx$3 AP.
\begin{table*}
\centering
\begin{tabular}{llcccccccccc}
\toprule
 & &\multicolumn{5}{c}{Mask} &\multicolumn{5}{c}{Bbox}\\
    \cmidrule(lr){3-7}                  
    \cmidrule(lr){8-12}   
Model & Backbone &AP &AP\textsubscript{50} &AP\textsubscript{S} &AP\textsubscript{M} &AP\textsubscript{L} &AP &AP\textsubscript{50} &AP\textsubscript{S} &AP\textsubscript{M} &AP\textsubscript{L}\\
\midrule
{Mask R-CNN} &{RES-50 + FPN} &18.2 & 30.3 & 4.6 & 15.0 & 25.7 & 23.3 & 34.4 & 14.0 & 23.5 & 27.4 \\
{Mask R-CNN w/SCP} &{RES-50 + FPN}& 25.3 & 36.3 & 12.4 & 23.0 & 34.4 & 25.7 & 36.6 & 19.1 & 27.6 & 30.6 \\
{Mask2Former} &{RES-50 + FPN} & 28.4 & 36.4 & 13.0 & 25.7 & 38.6 & 27.7 & 35.0 & 16.7 & 27.2 & 35.1 \\
{Mask2Former w/SCP} &{RES-50 + FPN} & 30.3 & 39.4 & 12.6 & 27.0 & 39.5 & 30.1 & 38.2 & 14.0 & 30.1 & 35.5 \\
\textbf{HAISTA-NET} &{RES-50 + FPN} & 51.2 & 68.5 & 41.8 & 53.5 & 57.2 & 52.1 & 69.3 & 51.4 & 56.8 & 53.1 \\
\textbf{HAISTA-NET w/SCP} &{RES-50 + FPN} & 51.2 & 68.9 & 41.4 & 53.5 & 57.7 & 53.3 & 69.2 & 53.4 & 58.4 & 54.1 \\
\textbf{HAISTA-NET} &{RES-101 + FPN} & 51.0 & 69.1 & 40.0 & 52.7 & 60.5 & 50.5 & 69.9 & 45.2 & 55.5 & 53.9\\
\textbf{HAISTA-NET w/SCP} &{RES-101 + FPN} & 52.2 & 69.8 & 41.6 & 54.4 & 59.2 & 54.5 & 70.0 & 54.5 & 59.9 & 55.6\\
\textbf{HAISTA-NET} &{RES-152 + FPN} & 52.8 & 71.0 & 39.8 & 55.4 & 60.1 & 52.6 & 71.7 & 49.6 & 57.3 & 54.6\\
\textbf{HAISTA-NET w/SCP} &{RES-152 + FPN} & 53.1 & 71.3 & 45.2 & 56.6 & 59.4 & 55.3 & 71.7 & 55.4 & 61.5 & 54.4\\
\textbf{HAISTA-NET} &{X-101 + FPN} & 52.0 & 70.9 & 40.1 & 54.9 & 58.1 & 51.5 & 71.4 & 45.3 & 57.5 & 52.7\\
$\textbf{HAISTA-NET w/SCP}$ & {X-101 + FPN} & $\textbf{54.9}$ & $\textbf{73.8}$ & $\textbf{47.8}$ & $\textbf{58.4}$ & $\textbf{61.5}$ & $\textbf{56.8}$ & $\textbf{74.4}$ & $\textbf{56.0}$ & $\textbf{62.3}$ & $\textbf{57.2}$\\
\bottomrule
\end{tabular}
\caption{The results of the models with different backbones and augmentation techniques. SCP represents the Simple Copy Paste data augmentation \cite{simplecopy} technique. We train HAISTA-NET 26 Epochs without Simple Copy Paste and 50 epochs with Simple Copy Paste. The training combinations of Mask R-CNN are same as HAISTA-NET. However, both versions of Mask2Former are trained with 25 epochs.}
\label{tab:genresults}
\end{table*}
\subsection{Training Parameters}
\label{sec:TrainingParameters}
We fine-tuned HAISTA-NET with 26 epochs for both ResNet \cite{resnet} and ResNeXt \cite{resnext} backbones using SGD optimizer. The learning rate of the optimizer was 0.0025. We set multi-step learning rate scheduling to decrease the learning rate by 10 at the 16th and 22nd epochs. In addition, we set a weight decay of 0.0001 and momentum of 0.9. 

We trained these three different models on the PSOB dataset in order to compare their performances. Since the PSOB dataset contains Human Attention Maps, RGB images, and annotation features such as category, bounding box, and mask polygon, we can apply the same process to these 3 models. However, only HAISTA-NET uses Human Attention Maps due to model configuration. Mask R-CNN and Mask2Former only use RGB images to train.

Since we are presenting a new dataset, PSOB, we must analyze the performance of other popular models such as Strong Mask R-CNN and Mask2Former on this dataset. For Mask R-CNN training, we used the same hyperparameters as HAISTA-NET because our model is based on MASK R-CNN. However, since Mask2Former is a transformer-based architecture, we tuned this model with 25 epochs, and the AdamW optimizer was used with a learning rate of 0.0025. Also, we set a weight decay of 0.05, epsilon of 1e-08, and betas of (0.9, 0.999)

We trained these three different models on the PSOB dataset in order to compare their performances. Since the PSOB dataset contains human attention maps, RGB images, as well as annotation features such as category, bounding box, and mask polygon, we can apply the same process to these three models. However, only HAISTA-NET uses human attention maps as a result of its model configuration. Mask R-CNN and Mask2Former only use RGB images to train.

PyTorch Framework is used for all implementations and configurations. In all experiments for training, we choose NVIDIA RTX 3090 TI single GPU with a batch size of 2. The image shuffling technique randomly selects images from the dataset in the training phase. We use two subprocesses for data loading in training.
\subsection{Inference}
\label{sec:Inference}
We performed inference for 4,118 annotations after the training. This was conducted using NVIDIA RTX 3090 TI with a single GPU with a batch size of 1. We obtained AP \cite{coco},
AP$\textsubscript{50}$, AP$\textsubscript{S}$, AP$\textsubscript{M}$, and AP$\textsubscript{L}$ metrics for all models. HAISTANET, which has strong data augmentation \cite{simplecopy} and a ResNeXt-101-FPN backbone, results in the best object mask and bounding box predictions.

In order to interact with images during inference time, we present a user-friendly graphical user interface (Figure~\ref{fig:interface}). With the help of the interface, users may upload an image in seconds and create hand-drawn sketches to use as a human attention map. Following the sketch, a user can observe object mask results as well as the bounding box of the object. Model guidance is possible by using the interface in one of two ways. In the first case, the image goes into the Mask R-CNN model for the user to see the results and to create a human attention map. In this case, a user may detect the missing object parts. Users may use this as a reference to create more precise partial sketches for HAISTA-NET. In the second case, the image is not fed into the Mask R-CNN model, and the user creates heuristic sketches without any prior knowledge of the segmentation results. This leaves the HAISTA-NET to make any predictions concerning the human attention map.
\section{Experiment}
In order to describe the performance of the models, we report our experimental results, which include AP metrics, mask/bounding box predictions, and analysis of the PSOB dataset. Additionally, we present relational analyses of the user’s sketch characteristics along with the curvature types and object scales used in the creation of the PSOB dataset.
\subsection{Main Results}
\label{sec:MainResults}
We evaluated our HAISTA-NET model with 4,118 test annotations of the PSOB dataset. Our results point out significant improvement over state-of-art models such as Strong Mask R-CNN and Mask2Former. Our analysis demonstrates apparent progress on mask precision deficiencies, such as object boundary outflow or under-covering. Results exhibit the effectiveness of a human attention map to overcome the mask prediction issues that standard models face with small-scale and high-curvature objects. Our experiments support the potential of our architecture for acquiring high precision segmentation masks (Figure~\ref{fig:predictions}). We demonstrate that HAISTA-NET produces better results than other models in all AP, AP$\textsubscript{50}$, A$\textsubscript{S}$, AP$\textsubscript{M}$, and AP$\textsubscript{L}$ results. Comparing the Res-Net-50FPN, Res-Net-101-FPN, Res-Net-152-FPN, and Res-NeXt101-FPN backbones, we achieved the best performance with Res-NeXt-101-FPN. (see Table~\ref{tab:genresults}). We also report that AP values improve if our model uses the simple copy-paste data augmentation method.

We benchmark the HAISTA-NET Strong Mask R-CNN baseline by utilizing output images. HAISTA-NET demonstrates better results with challenging high-curvature objects and small-scale objects (see Table~\ref{tab:curveresults}).
\subsection{Multiple Factor Analysis}
\label{sec:MultipleFactorAnalysis}
Using multiple factor analysis \cite{anova1}, we classified the object given in test data according to scale, curvature, and level of assistance. In order to perform the analysis, we first ran the inference with HAISTA-NET. Later on, we retrieved the mask predictions and split results into classified partitions such as small-scale objects with low-curvature and major user assistance, and large-scale objects with medium curvature and minor user assistance.

\begin{figure}[htbp]
\centering%
\includegraphics[width=.9\linewidth]{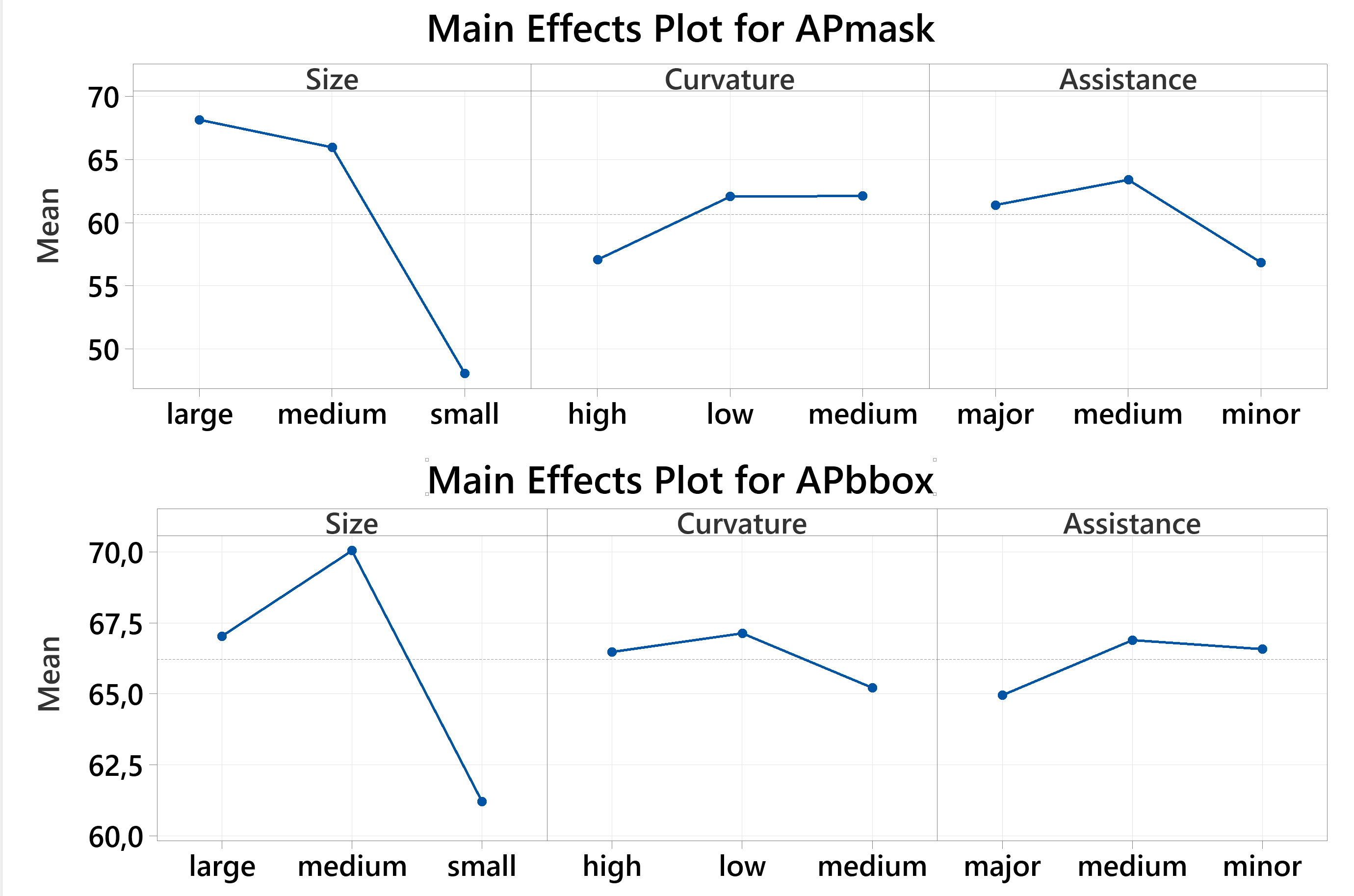}
\includegraphics[width=.9\linewidth]{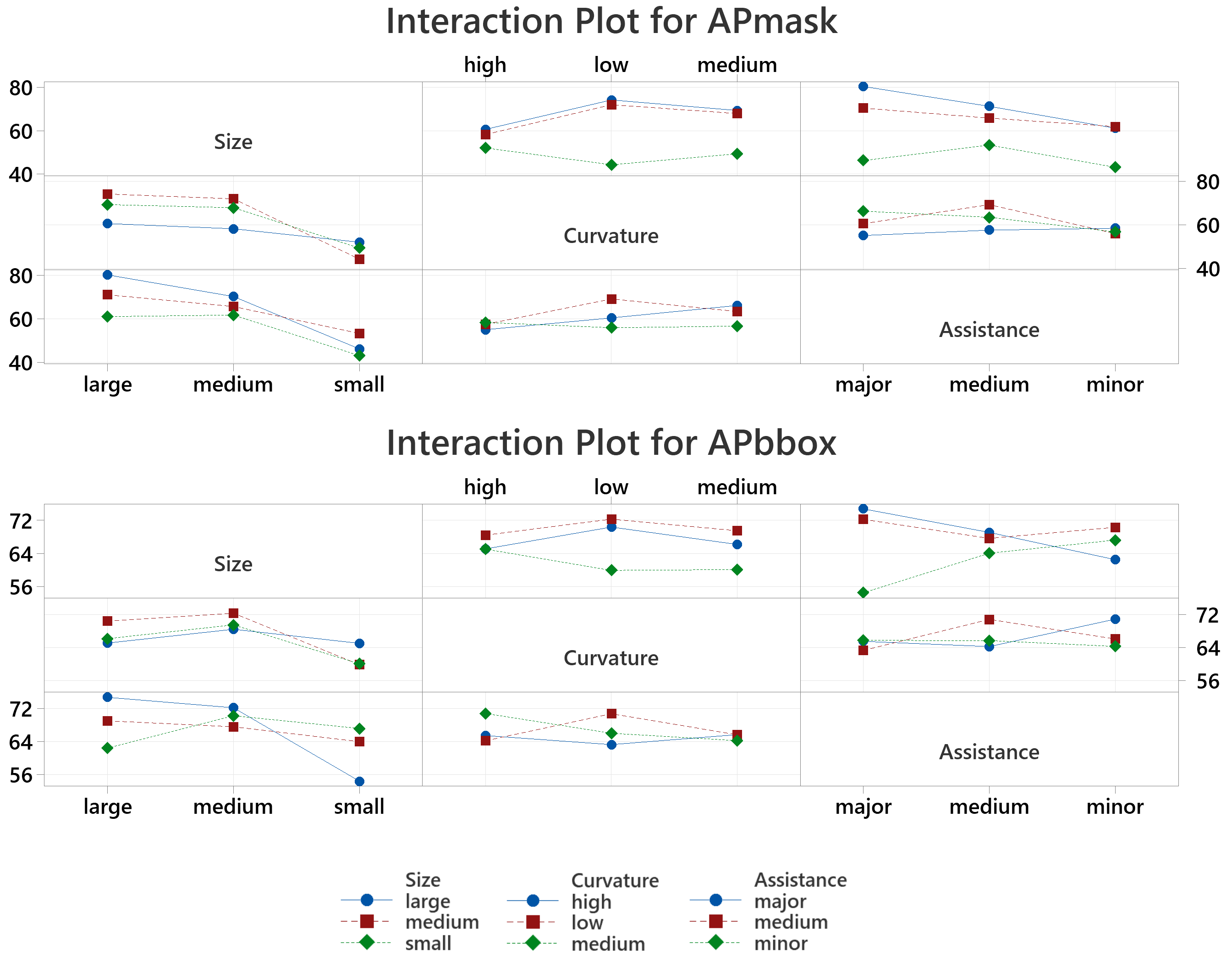}
   \caption{The graphs demonstrate the main and interaction effects of AP$\textsubscript{Mask}$ and AP$\textsubscript{Bbox}$ values generated using HAISTA-NET.}
\label{fig:anova}
\end{figure}

We conducted a factorial analysis of variance (ANOVA, two-way) to compare the main effects on size, curvature, and user assistance. We also compared the correlated interaction effects on size $\otimes$ curvature, size $\otimes$ assistance, and curvature $\otimes$ assistance. A statistically significant difference exists between size, curvature, and assistance at p-value \((p) < 0.05\). The main effect of size (\(F\) (2,5) = 123.59, \(p\) = 0.001), curvature (\(F\) (2,5) = 9.13, \(p\) = 0.021), and assistance (\(F\) (2,5) = 9.21, \(p\) = 0.021) in AP$\textsubscript{Mask}$ are significant such that objects having large size, low curvature, and major or medium assistance receive higher scores. In AP$\textsubscript{Mask}$ analysis, there is a significant interaction between size and curvature (\(F\) (4,5) = 14.05, \(p\) = 0.006) as well as curvature and assistance (\(F\) (4,5) = 5.28, \(p\) = 0.048). The main effect of size (\(F\) (2,5) = 16.04, \(p\) = 0.007), curvature (\(F\) (2,5) = 6.23, \(p\) = 0.044), and assistance (\(F\) (2,5) = 9.87, \(p\) = 0.018) in AP$\textsubscript{Bbox}$are significant such that objects with a large or medium size, high or low curvature, and minor or medium assistance received higher scores. For AP$\textsubscript{Bbox}$ analysis, there is a significant interaction between size and curvature (\(F\) (4,5) = 15.82, \(p\) = 0.005)), size and assistance (\(F\) (4,5) = 10.54, \(p\) = 0.012), as well as curvature and assistance (\(F\) (4,5) = 7.16, \(p\) = 0.027) (Figure~\ref{fig:anova}).
\subsection{Curvature-Based Average Precision}
\label{sec:Curvature-BasedAveragePrecision}
We report the results of the average precision (AP) values of objects in all scales according to curvature classification. In order to analyze the results of curvature-based AP, we divide the objects of the PSOB test set into low curvature, medium curvature, and high curvature types. In addition, we compare HAISTA-NET AP results with Strong Mask R-CNN for AP$\textsubscript{low-curvature}$, AP$\textsubscript{medium-curvature}$ and AP$\textsubscript{high-curvature}$, demonstrating that HAISTA-NET is more successful than Strong Mask R-CNN for all curvature types (see Table~\ref{tab:curveresults}).
\begin{table}[htbp]
\begin{center}
\begin{tabular}{lccc}
\toprule
&\multicolumn{3}{c}{Mask}\\
\cmidrule(lr){2-4} 
Model &AP$\textsubscript{low-curvature}$& AP$\textsubscript{med.-curvature}$ &AP$\textsubscript{high-curvature}$\\
\midrule
${M}$ &25.5 &21.4 &21.4\\
\textbf{${H}$$\textsubscript{1}$}  &58.1 &53.3 &50.4\\
\textbf{${H}$$\textsubscript{2}$} &\textbf{61.2} &\textbf{58.2} &\textbf{51.9}\\
\bottomrule
\end{tabular}
\caption{Results of the AP values of the models in different curvature types. ${M}$ is Mask R-CNN w/SCP + Res-50-FPN, ${H}$$\textsubscript{1}$ is HAISTA-NET w/SCP + Res-50-FPN, and ${H}$$\textsubscript{2}$ is HAISTA-NET w/SCP + X-101-FPN}
\label{tab:curveresults}
\end{center}
\end{table}
\subsection{PSOB Interaction Time Analysis}
\label{sec:DataAnnotationAnalysis}
We annotated 18,677 objects according to 115 categories and three different scales (small, medium, and large). This data was collected from 30 users. We stored critical numerical metrics such as total interaction time, area, perimeter (length) of segmentation polygon (mask), length of sketch input, percentage of covering object boundaries with the sketch, stroke count, and sketching time except latency. One of the most crucial factors of this analysis is the time benchmark. Since our model takes extra user input during training and inference, shorter interaction is better. We developed sketch-based interaction considering this situation. Time analysis shows that total interaction time is roughly three times longer than sketching time because interaction time has latency. The human reasoning process causes an extension of time for detecting an object’s exact boundaries. In addition, we performed multiple regression \cite{regression1} analysis to determine which factors affect total interaction time. We set the total interaction time as a response (dependent variable) and the object’s curvature number, length of segmentation polygon, stroke count, and length of sketch input as predictors (independent variables). As a result, all the independent variables significantly predict total interaction time (Table ~\ref{tab:annanalysis}) because the \(p\) values of all independent variables are less than 0.05. Furthermore, this regression model is statistically significant with R$\textsuperscript{2}$ of 76.27\%.

\begin{table}[htbp]
    \centering
    \begin{tabular}{lccc}
\toprule
    &Train & Validation & Test\\
\midrule
Interaction Time [sec] & 7.2 & 8.3 & 8.4 \\ 
        Sketching Time [sec] & 2.0 & 3.2 & 3.3 \\ 
        No. Of Curvatures & 7.6 & 7.5 & 8.0  \\ 
        Perimeter of Polygon & 695.4 & 679.6 & 688.2 \\
        LS/PP (\%) & 19.9 & 31.2 & 31.2 \\ 
        Stroke Count & 6.5 & 6.2 & 6.0 \\ 
        Annotation Count & 10450 & 4109 & 4118 \\ 
\bottomrule
    \end{tabular}
    \caption{PSOB dataset stores the pieces of information about sketched objects. LS/PP represents "Length of Sketch" over "Perimeter of Polygon." This ratio represents the number of pixels that are drawn on the object's boundaries as a percentage.}
    \label{tab:annanalysis}
\end{table}
\subsection{User Study}
\label{sec:UserStudy}
Users can annotate objects manually if the model output is not satisfying. However, manual annotation is time-consuming compared with PSOB’s partial annotation. In particular, retrieving high-detail object polygons such as LVIS \cite{lvis} annotations requires extra attention and focus. Since LVIS contains fine-grained polygons of object boundaries, we were curious about the interaction time needed for annotating an object’s mask with hand-drawn sketches. In order to perform this experiment, we collected 643 objects of different sizes (small, medium, and large) to analyze the average interaction time and mIOU values of LVIS-Like sketches and rough sketches. We then compared the results with the PSOB average interaction time (Table ~\ref{tab:userstudy}) and mIOU value (Table~\ref{tab:userstudy2}). The following steps are specified for performing the study:
\begin{itemize}
\item Users draw the boundaries of ground truth masks with fine-grained characteristics such as LVIS annotations, and interaction time [sec] is recorded.
\item Users draw the boundaries of ground truth masks as a rough sketch and interaction time [sec] is recorded.
\item We feed the model with a human attention map from the PSOB dataset to determine how close our model results are to the LVIS mask.
\end{itemize}
Results can be summarized as follows; 
\begin{itemize}
\item LVIS-Like interactions have the best quality but take the longest time and require the greatest effort. 
\item Rough sketches are fast but do not cover the curvatures of an object due to its characteristics. 
\item Since PSOB annotations are partial data, they take the shortest time. As a result, we acquire mask qualities closest to LVIS-Like, without complete manual annotation, when we feed the model with the human attention map.
\end{itemize}
\begin{table}[htbp]
\begin{center}
\begin{tabular}{lcccc}
\toprule
\multicolumn{2}{c}{Object}&\multicolumn{3}{c}{Average Sketch Time [sec]}\\
\cmidrule(lr){1-2} 
\cmidrule(lr){3-5} 
Scale & Number &LVIS-Like& Rough &PSOB\\
\midrule
small &27 &257 &15 &9\\
medium &199 &116 &17 &8\\
large &238 &117 &16 &9\\
\bottomrule
\end{tabular}
\caption{Results of the average sketch time differences of LVIS-Like, Rough, and PSOB sketches for different scale objects.}
\label{tab:userstudy}
\end{center}
\end{table}
\begin{table}[htbp]
\begin{center}
\begin{tabular}{lccc}
\toprule
&\multicolumn{3}{c}{mIOU}\\
\cmidrule(lr){2-4} 
Interaction Type &Small& Medium &Large\\
\midrule
LVIS-Like &99$\%$ &97$\%$ &96$\%$\\
Rough  &91$\%$ &82$\%$ &78$\%$\\
PSOB &98$\%$ &93$\%$ &91$\%$\\
\bottomrule
\end{tabular}
\caption{Results of the mean intersection over union (mIOU) values of LVIS-Like, Rough, and PSOB sketches for different scale objects.}
\label{tab:userstudy2}
\end{center}
\end{table}
\section{Conclusion}
Finding the correct boundaries for small-scale objects and objects with high curvature points is a challenging task in instance segmentation. In this study, we present a novel approach, HAISTA-NET, and the concept of human attention maps, which are shown to achieve significant improvements in these areas compared with current fully automated state-of-the-art algorithms. Our method requires a minimal amount of input from users and does not require longer training and inference time in any noticeable manner. Moreover, we also present a new dataset, PSOB (Partial Sketch Object Boundaries), that combines human attention maps with the LVIS dataset for instance segmentation. Our user-friendly interface brings a new perspective to annotation and interaction techniques. We also provide an extensive analysis of the factors that affect the performance of our architecture and state-of-the-art methods. Multiple factor analysis, according to AP$\textsubscript{Bbox}$and AP$\textsubscript{Mask}$ metrics, shows the most affecting factors.

HAISTA-NET is easy to use and can be extended to other
computer vision tasks such as object detection, panoptic segmentation, etc. It can be easily implemented using other visual software applications that use CNNs with a minimal amount of user input. We hope that the evidence we provide in this study encourages other authors to incorporate HAISTA-NET in their own research to improve performance.
{\small
\bibliographystyle{ieee_fullname}
\bibliography{egbib}
}

\end{document}